\documentclass[conference,10pt]{IEEEtran}
\IEEEoverridecommandlockouts
\usepackage{amsmath,amssymb,amsfonts}
\usepackage{algorithmic}
\usepackage{graphicx}
\usepackage{tabularx}
\usepackage{booktabs}
\usepackage{microtype}
\usepackage{multirow} 
\usepackage{textcomp}
\usepackage{makecell} 
\usepackage{xcolor}
\usepackage{colortbl}
\definecolor{lightpurple}{RGB}{112, 48, 160}
\def\BibTeX{{\rm B\kern-.05em{\sc i\kern-.025em b}\kern-.08em
    T\kern-.1667em\lower.7ex\hbox{E}\kern-.125emX}}
\usepackage[numbers]{natbib}
\usepackage{hyperref}

\begin{document}

\title{\includegraphics[width=.8cm]{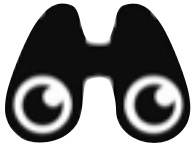} Language Models Can See Better: 
    Visual Contrastive Decoding For LLM Multimodal Reasoning \vspace{-.3em}
    
\thanks{$^{\star}$Corresponding author. $^{\dagger}$Work done during Haoqin Tu's stay at UCAS, China. This work was supported by NSFC under 62272456.}
\thanks{© 2025 IEEE.  Personal use of this material is permitted. Permission from IEEE must be obtained for all other uses, in any current or future media, including reprinting/republishing this material for advertising or promotional purposes, creating new collective works, for resale or redistribution to servers or lists, or reuse of any copyrighted component of this work in other works.}
}

\author{
\IEEEauthorblockN{Yuqi Pang$^{1,2}$\quad Bowen Yang$^{1,2}$\quad Haoqin Tu$^{3\dagger}$\quad Yun Cao$^{1,2,\star}$\quad Zeyu Zhang$^{1,2}$}
\IEEEauthorblockA{$^1$ Institute of Information Engineering, Chinese Academy of Sciences, Beijing, China}
\IEEEauthorblockA{$^2$ School of Cyber Security, University of Chinese Academy of Sciences, Beijing, China}
\IEEEauthorblockA{$^3$ Computer and Science Engineering, University of California, Santa Cruz, CA, USA}
\IEEEauthorblockA{\{pangyuqi, yangbowen\}@iie.ac.cn \quad 
 tuisaac163@gmail.com \quad \{caoyun, zhangzeyu\}@iie.ac.cn}
}
\maketitle
\begin{abstract}

Although Large Language Models (LLMs) excel in reasoning and generation for language tasks, they are not specifically designed for multimodal challenges. Training Multimodal Large Language Models (MLLMs), however, is resource-
intensive and constrained by various training limitations. 
In this paper, we propose the Modular-based Visual Contrastive Decoding (MVCD) framework to move this obstacle. Our framework leverages LLMs' In-Context Learning (ICL) capability and the proposed visual contrastive-example decoding (CED), specifically tailored for this framework, without requiring any additional training. 
By converting visual signals into text and focusing on contrastive output distributions during decoding, we can highlight the new information introduced by contextual examples, explore their connections, and avoid over-reliance on prior encoded knowledge.
MVCD  enhances LLMs’
visual perception to make it see and reason over the input visuals.
To demonstrate MVCD's effectiveness, we conduct experiments with four LLMs across five question answering datasets. Our results not only show consistent improvement in model accuracy but well explain the effective components inside our decoding strategy. Our code will be available at \url{https://github.com/Pbhgit/MVCD}.
\end{abstract}
\section{Introduction}
\label{sec:intro}
Large Language Models (LLMs)~\cite{r4,r5} have made significant advances in Natural Language Processing (NLP), demonstrating extraordinary capabilities such as instruction following~\cite{r6}, In-Context Learning (ICL)~\cite{ic}, and Chain-of-Thought (CoT)~\cite{COT} reasoning by scaling up both data and model size. The introduction of vision language models has further enhanced these capabilities by enabling reasoning over visual content in diverse vision language tasks, such as visual question answering (VQA)~\cite{liu2024visual,tu2023sight}, visual dialogue~\cite{tu2023resee}. 
Yet, what is the best representation, in efficiency and performance, for transferring LLMs to Multimodal Large Language Models (MLLMs) remains an open question. 
Transforming LLMs to MLLMs typically requires combining visual and textual signals using extra training parameters and the corresponding data~\cite{r1,r2,r3, zhang2023video,tu2023sight}.  This paradigm not only demands significant computational
resources and extensive data, but also requires model retraining whenever a new modality is introduced. 

\begin{figure*}[htb]
\centering
\makebox[\textwidth]{\includegraphics[width=0.95\linewidth]{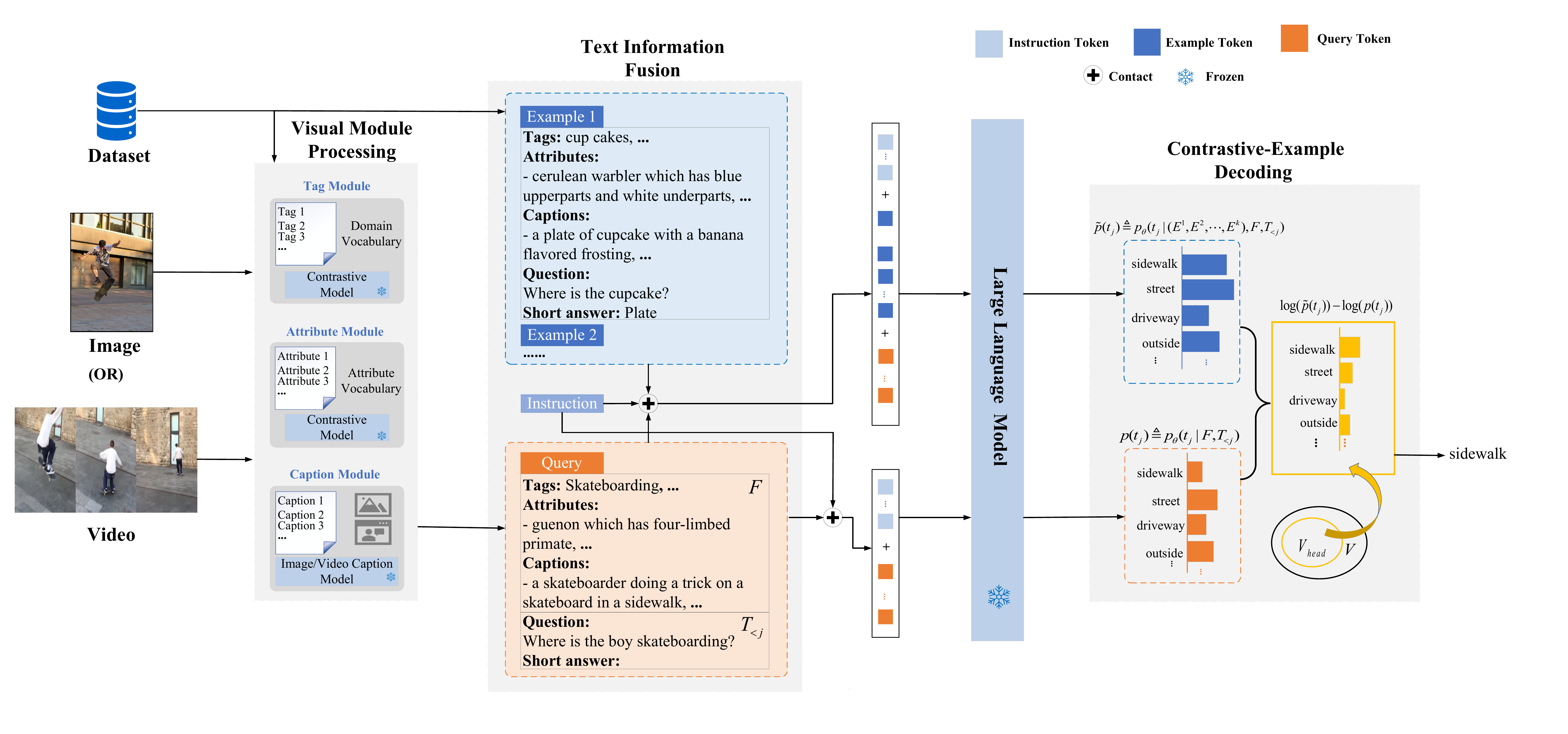}}
\vspace{-1em}
\caption{Our MVCD workflow contains three main components: Visual Module Processing, Text Information Fusion and Contrastive-Example Decoding.}
\label{fig:frame}
\end{figure*}

Existing efficient multimodal methods focus on integrating probability-based visual control to the decoding stage~\cite{su2022language,tu2023zerogen} or simply bridging visual and language modules using symbolic representations~\cite{r9,wei2024diffusion}. 
These methods acquire multimodal knowledge without task-specific training, thereby minimizing the computational cost of multimodal reasoning using LLM. Although efficient MLLM methods reduce resource requirements through lightweight designs, they often underperform on complex tasks like multimodal reasoning
and generation. These models often prioritize knowledge from a single modality during multimodal reasoning, neglecting the need to explore connections from different modalities. As a result, they fail to fully capture patterns within these data pairs, leading to suboptimal performance in related tasks and exposing potential issues while deploying these models~\cite{tu2024many,leevhelm}.

To overcome those shortcomings, we equip LLMs with visual reasoning capability and propose a plug-and-play framework, MVCD. As shown in Fig.~\ref{fig:frame}, adding a visual perception module to the LLM to extract text features (e.g., tags, attributes, and captions) enables task handling without retraining. To achieve multimodal reasoning, we are aware of the need to capture connections between different modal data pairs to guide response. Specifically, we first propose the use of the text-only ICL format to better instruct MLLMs for reasoning tasks. Based on traditional contrastive decoding~\cite{r7,r22}, we proposed contrastive-example decoding (CED). By strategically selecting appropriate contextual examples, CED compares output probability distributions with and without these examples, filtering low-probability tokens under adaptive constraints to optimize reasoning and address the limitations of prior knowledge.
We apply MVCD to five question answering datasets, showing that CED enhances LLMs' visual perception and robustness across different shot settings while preserving their original strengths.
In summary, our main contributions are: \textbf{(1) }We explore multimodal reasoning with LLMs and propose the MVCD framework, which integrates a visual perception module into an LLM to enhance its `seeing' capability without the need for additional training. \textbf{(2) }We introduce an adaptive contrastive-example decoding method, specifically designed for the MVCD framework, to enhance LLMs’ comprehension and reasoning in complex visual scenarios. 
\textbf{(3) }Extensive experiments on five datasets validate MVCD’s efficacy and explore the impact of model configurations, including multimodal shot count, example selection strategy, and reasoning components.

\section{Methodology}
\label{sec:method}
\subsection{Problem Definition}
\label{sec:problem}
For visual perception tasks, we formally define it as: given a visual-text dataset, such as image-text pairs  $S=\{(I^1,T^1),(I^1,T^1),\cdots,(I^n,T^n)\}$ (or video-text pairs $S=\{(V^1,T^1),(V^1,T^1),\cdots,(V^n,T^n)\}$), and given an image $I$ (or a video $V$), a language model $\theta$ (the weights of the LLM),  and input text 
$T=\{t_1,t_2,\cdots,t_{j-1}\}$, where $t_i$ is the $i$-$th$  token, our goal is to generate the corresponding output text.

As shown in Fig.~\ref{fig:frame}, the visual perception module converts images or videos to text, such as tags, attributes, and captions. For the current image $i$, at decoding step $j$, the probability output of the original LLMs based on all textual information of image $i$ is $p(t_j)$. After prepending contextual examples information to the text, the output probability of $t_j$ becomes $\tilde{p}(t_j)$.  
It is important to note that for the prediction of $t_j$, the token in the contrastive probability output space should be dynamically selected based on the decoding strategy of the language model.
\subsection{Visual Module}
\label{sec:modular}
\paragraph{Image Modularization} Inspired by the LENS \cite{r9} feature extraction method, we match each image with the corresponding tags, attributes, and captions. \textbf{Tag/Attribute Module -- } We use CLIP  \cite{r10} (i.e., OpenCLIP-H/14 and CLIP-L/14) as our default visual encoder to obtain $N$ tags for each image and, with task-specific prompts from \cite{r11}, obtain $N$ attributes.
\textbf{Image Caption Module -- }We adopt BLIP \cite{r12} image caption model to generate $N$ captions for each image using random top-$k$ sampling \cite{r13}, where $k$ represents the desired number of captions. 
\paragraph{Video Modularization}
\textbf{Tag/Attribute Module --} To obtain high-quality visual-language representations, we employ the recently released LanguageBind \cite{r14} to extract features and obtain $N$ tags/attributes for each video. \textbf{Video Caption Module -- }We select Video-LLaVA \cite{r15} as the video language model for
captioning, adjusting the top $k=50$ and temperature parameters to generate $N$ diverse captions.
\begin{table*}[htbp]
\centering
\renewcommand\arraystretch{0.8}
\scriptsize
\caption{Comparison of LLMs combined with CED (MVCD) and the Dola with LENS in zero-, one-, three-, and five-shot settings.}
\vspace{-0.5em}
\begin{tabular}{@{}l|l|ccccc|*{5}{c}@{}}
\toprule
LLM & Method & VQAv2 & OKVQA & MME & MSRVTT-QA &MSVD-QA & VQAv2 & OKVQA & MME & MSRVTT-QA &MSVD-QA\\ 
\midrule
\multicolumn{2}{c|}{}&\multicolumn{5}{c|}{Original (Zero-shot)}&  \multicolumn{5}{c}{One-shot} \\ \midrule

\multirow{3}{*}{LLaMA2-7B}  &LENS & 41.80 & 25.27 & 1341 & 42.17 & 63.93 &51.52 & 29.97 & 1402 & 42.97 & 65.31 \\
        &Dola & 43.42 & 25.00&  1332 & 42.25 & 62.08&\textcolor{blue}{+0.61} & \textcolor{blue}{+1.06} &  \textcolor{red}{-12} &  \textcolor{blue}{+1.48} & \textcolor{blue}{+0.85}        \\
        &MVCD&-&-&-&-&-&\textcolor{blue}{+0.74} & \textcolor{blue}{+0.03} & \textcolor{red}{-21} & \textcolor{blue}{+0.09} & \textcolor{red}{-0.41} \\
\midrule
\multirow{3}{*}{LLaMA2-13B} &LENS &46.32 & 22.39 & 1233 & 42.86 & 66.53& 52.09 & 23.48 & 1265 & 44.97 & 65.71 \\
        &Dola &49.14 &  26.70 &1305 & 43.65& 64.80& \textcolor{blue}{+1.31} &  \textcolor{blue}{+4.32} & \textcolor{red}{-8} & \textcolor{blue}{+0.70} &  \textcolor{blue}{+1.41} \\
        &MVCD   &-&-&-&-&-&\textcolor{blue}{+0.02} &  \textcolor{blue}{+4.63} &\textcolor{red}{-45} &  \textcolor{blue}{+0.81} & \textcolor{red}{-0.40} \\
\midrule
\multirow{3}{*}{LLaMA3-8B}  &LENS  &46.15 & 25.93 & 1508 & 41.10 & 55.88& 53.83 & 28.95 & 1523 & 42.57 & 57.96      \\
            &Dola &45.09 & 28.60& 1514&  42.93& 60.72&\textcolor{blue}{+0.84} & \textcolor{blue}{+2.69} & \textcolor{blue}{+4} & \textcolor{blue}{+0.09} & \textcolor{blue}{+3.61}\\
            &MVCD &-&-&-&-&-&\textcolor{blue}{+1.07} & \textcolor{blue}{+0.65} & \textcolor{blue}{+15} & \textcolor{red}{-1.12} & \textcolor{blue}{+3.67} \\
\midrule
\multirow{3}{*}{Qwen2-7B}   &LENS &49.39 & 29.49 & 1466 & 41.79 & 64.49& 54.18 & 29.71 & 1498 & 43.12 & 67.90 \\
            & Dola & 48.99 & 28.81&  1489& 42.77 &64.98&\textcolor{blue}{+1.08} & \textcolor{blue}{+1.36} &  \textcolor{blue}{+10} & \textcolor{blue}{+0.57} & \textcolor{blue}{+2.10} \\
            &MVCD   &-&-&-&-&-&\textcolor{blue}{+1.72} & \textcolor{blue}{+2.79} & \textcolor{blue}{+71} & \textcolor{blue}{+0.60} & \textcolor{blue}{+1.55} \\
\midrule
\multicolumn{2}{c|}{}&\multicolumn{5}{c|}{Three-shot} &\multicolumn{5}{c}{Five-shot} \\ \midrule

\multirow{3}{*}{LLaMA2-7B}  &LENS & 53.06 & 31.60 & 1441& 46.08 & 68.98 & 52.43 & 31.33 & 1472 & 46.08 & 68.98 \\
        &Dola & \textcolor{blue}{+2.80} & \textcolor{blue}{+0.70} &  \textcolor{blue}{+13} &  \textcolor{blue}{+1.27} & \textcolor{blue}{+1.39}  &\textcolor{blue}{+3.01} & \textcolor{red}{ -0.63} &  \textcolor{red}{-43} &  \textcolor{blue}{+0.20} & \textcolor{blue}{+2.46}          \\
           &MVCD & \textcolor{blue}{+2.97} & \textcolor{blue}{+0.40} & \textcolor{blue}{+17} & \textcolor{blue}{+1.80} & \textcolor{blue}{+0.82} & \textcolor{blue}{+2.61} & \textcolor{blue}{+0.87} & \textcolor{blue}{+11} & \textcolor{blue}{+0.19} & \textcolor{blue}{+1.63}\\
\midrule
\multirow{3}{*}{LLaMA2-13B} &LENS & 58.32 & 34.80 & 1481 & 46.79 & 72.65& 56.93 & 35.18 & 1562 & 47.48 & 71.43 \\
        &Dola & \textcolor{blue}{+2.54} &  \textcolor{red}{-0.20} & \textcolor{red}{-4} & \textcolor{blue}{+0.76} &  \textcolor{red}{ -0.07} & \textcolor{blue}{+2.47} &  \textcolor{blue}{+0.22} & \textcolor{red}{-14} & \textcolor{blue}{+2.71} &  \textcolor{blue}{+0.45} \\
        &MVCD  & \textcolor{blue}{+3.08} &  \textcolor{blue}{+4.20} &\textcolor{red}{-11} &  \textcolor{blue}{+1.27} & \textcolor{blue}{+0.41} & \textcolor{blue}{+3.40} &  \textcolor{blue}{+1.24} &\textcolor{red}{-42} &  \textcolor{blue}{+2.24} & \textcolor{blue}{+0.59}\\
\midrule
\multirow{3}{*}{LLaMA3-8B}  &LENS  & 56.67 & 32.11 & 1650 & 47.59 & 61.22     &57.44 & 30.91 & 1715 & 48.84 & 66.12\\
            &Dola &\textcolor{red}{ -0.66} & \textcolor{blue}{+5.22} & \textcolor{blue}{+22} & \textcolor{blue}{+0.43} & \textcolor{blue}{+5.07}&\textcolor{blue}{+2.26 } & \textcolor{blue}{+6.99} & \textcolor{red}{-22} & \textcolor{blue}{+0.39} & \textcolor{blue}{+1.64}\\
            &MVCD &\textcolor{blue}{+2.12} & \textcolor{blue}{+6.29} & \textcolor{blue}{+27} & \textcolor{blue}{+0.54} & \textcolor{blue}{+6.00}&\textcolor{blue}{+3.66} & \textcolor{blue}{+3.49} & \textcolor{blue}{+47} & \textcolor{blue}{+1.19} & \textcolor{blue}{+0.82} \\
\midrule
\multirow{3}{*}{Qwen2-7B}   &LENS & 58.26 & 33.60 & 1591 & 46.45 & 73.01 & 58.63 & 32.51 & 1641 & 48.13 & 75.31\\
            &Dola & \textcolor{blue}{+1.27} & \textcolor{red}{-0.89} &  \textcolor{red}{-8} & \textcolor{blue}{+1.03} & \textcolor{red}{-0.69} & \textcolor{blue}{+1.94} & \textcolor{blue}{ +2.32} &  \textcolor{blue}{+9} & \textcolor{blue}{+3.77} & \textcolor{blue}{+0.66}\\
            &MVCD   &\textcolor{blue}{+6.09} & \textcolor{blue}{+4.00} & \textcolor{blue}{+42} & \textcolor{blue}{+1.52} & \textcolor{blue}{+1.37}&\textcolor{blue}{+2.47} & \textcolor{blue}{+4.25} & \textcolor{blue}{+24} & \textcolor{blue}{+4.94} & \textcolor{blue}{+2.55} \\
\bottomrule
\end{tabular}
\label{tab:22}
\end{table*}
\subsection{Text Information Fusion}
\label{sec:fusion}  
To enhance multimodal information expression, we obtain descriptive textual features for every visual input (i.e., image or video) by integrating tags, attributes, and captions, providing complete example fusion as the foundation for the subsequent task. To fully leverage the potential of LLMs, we adopt ICL and introduce a visual-incorporated ICL method. This approach aims to enhance the way LLMs utilize visual information during the decoding process. 
We select $k$ suitable data pairs $(I^1, T^1), \cdots (I^k, T^k)$ from the dataset $S$ for the current image $I$, where $k \in \{1, 2, \cdots, n\}$. In all experiments, contextual examples were selected from $S$ based on the question type (matching the current question), rather than randomly.
Using the visual perception module described in section \ref{sec:modular}, $N$ tags, attributes, and captions are extracted for each image and concatenated as ``Tags: {Top-N tags}, Attributes: {Top-N attributes}, Captions: {Top-N captions}''. These are combined to form descriptive features. Let $F=(Tags,Attributes,Captions)$ and $F^i=(Tags^i,Attributes^i,Captions^i)$. The former represents the descriptive features of the image $I$, while the latter represents those of the $i$-$th$ image in the data pairs. The relevant textual information for each data pair is combined into $E^i=(F^i,T^i)$, representing the complete example information for the $i$-$th$ data pair. The combination of the $k$ examples $(E^1,E^2,\cdots,E^k)$ serves as contextual examples for the contrastive analysis in section \ref{sec:ccd}.
\begin{table}[t]
\centering
\caption{Comparison between MVCD using Qwen2-7B and MLLMs.}
\vspace{-0.5em}
{\setlength{\tabcolsep}{-0pt} 
\renewcommand\arraystretch{0.8}
\begin{minipage}{0.48\linewidth}
\centering
\resizebox{\linewidth}{!}{
\begin{tabular}{@{}lccc@{}}
\toprule
\multirow{2}{*}{MLLMs} & \multicolumn{3}{c}{Image Question Answering} \\
\cmidrule(lr){2-4} 
& VQAv2 \quad & OKVQA & MME \\
\midrule
\rowcolor{gray!10} Flamingo~\cite{r3} & 51.8 & 44.7 & - \\
\rowcolor{gray!10} BLIP-2~\cite{li2023blip} & 41.0 & - & 1293.8 \\
\rowcolor{gray!10} Qwen-VL~\cite{r1} & 78.2 & 56.6 & 1487.5 \\
\rowcolor{gray!10} LLaVA-v1.5~\cite{liu2024visual} & 78.5 & - & 1510.7 \\
\midrule
Ours (Three-shot) & 64.35 & 37.60 & 1633 \\
\bottomrule
\end{tabular}}
\end{minipage}
\hfill
\begin{minipage}{0.48\linewidth}
\centering
\resizebox{\linewidth}{!}{
\begin{tabular}{@{}lcc@{}}
\toprule  
\multirow{2}{*}{MLLMs} & \multicolumn{2}{c}{Video Question Answering} \\
\cmidrule(lr){2-3} 
& MSRVTT-QA & MSVD-QA \\
\midrule
\rowcolor{gray!10} LLaMA-Adapter~\cite{zhang2023llama} & 43.8 & 54.9 \\
\rowcolor{gray!10} Video-LLaMA~\cite{zhang2023video} & 29.6 & 51.6 \\
\rowcolor{gray!10} Video-ChatGPT~\cite{maaz2023video} & 49.3 & 64.9 \\
\rowcolor{gray!10} Instruct-BLIP~\cite{r2} & 22.1 & 41.8 \\
\midrule
Ours (Four-shot) & 52.11 & 79.59 \\
\bottomrule
\end{tabular}}
\end{minipage}
}
\label{tab:base}
\end{table}
\subsection{Contrastive-Example Decoding -- CED}
\label{sec:ccd}
When LLM predicts the $j$-th token $t_j$, the original decoding concatenates $F$ and $T$ into $(F,T_{<j})$ to predict the output probability under task-specific instructions. If we consider the hidden information contained in the contextual examples, it is prefixed to the above information with $k$ contextual examples to form the final input: $((E^1,E^2,\cdots,E^k),F,T_{<j})$. However, in cases where contextual examples contain knowledge that is out-of-distribution with respect to the language model $\theta$, which may cause the model to struggle with effectively focusing on the associations between examples and to overly rely on the prior knowledge encoded in $\theta$. 

To address this issue, we propose a decoding strategy, contrastive-example decoding (CED). Prior knowledge is extracted from the model's original distribution, with the prior knowledge modeling the original output probability distribution as $p(t_j)\triangleq p_{\theta}(t_{j}|F,T_{<j})$. 
Next, we use contextual examples to adjust the model's original output probability distribution to obtain the adjusted probability distribution: $\tilde{p}(t_j)\triangleq p_{\theta}(t_{j}|(E^1,E^2,\cdots,E^k),F,T_{<j})$, 
focusing more on the knowledge from the examples and their relationships. Essentially, outputs that become much more likely when contextual examples are included are preferred. Our goal is to enhance the adjusted output while minimizing reliance on incorrect elements of the original output:
\begin{equation}
\text{log}\tilde{p}(t_j)-\text{log}p(t_j),
\label{eq4}
\end{equation}

\begin{figure}[t]
\begin{minipage}[b]{.458\linewidth}
  \centering
  \centerline{\includegraphics[width=1.1\linewidth]{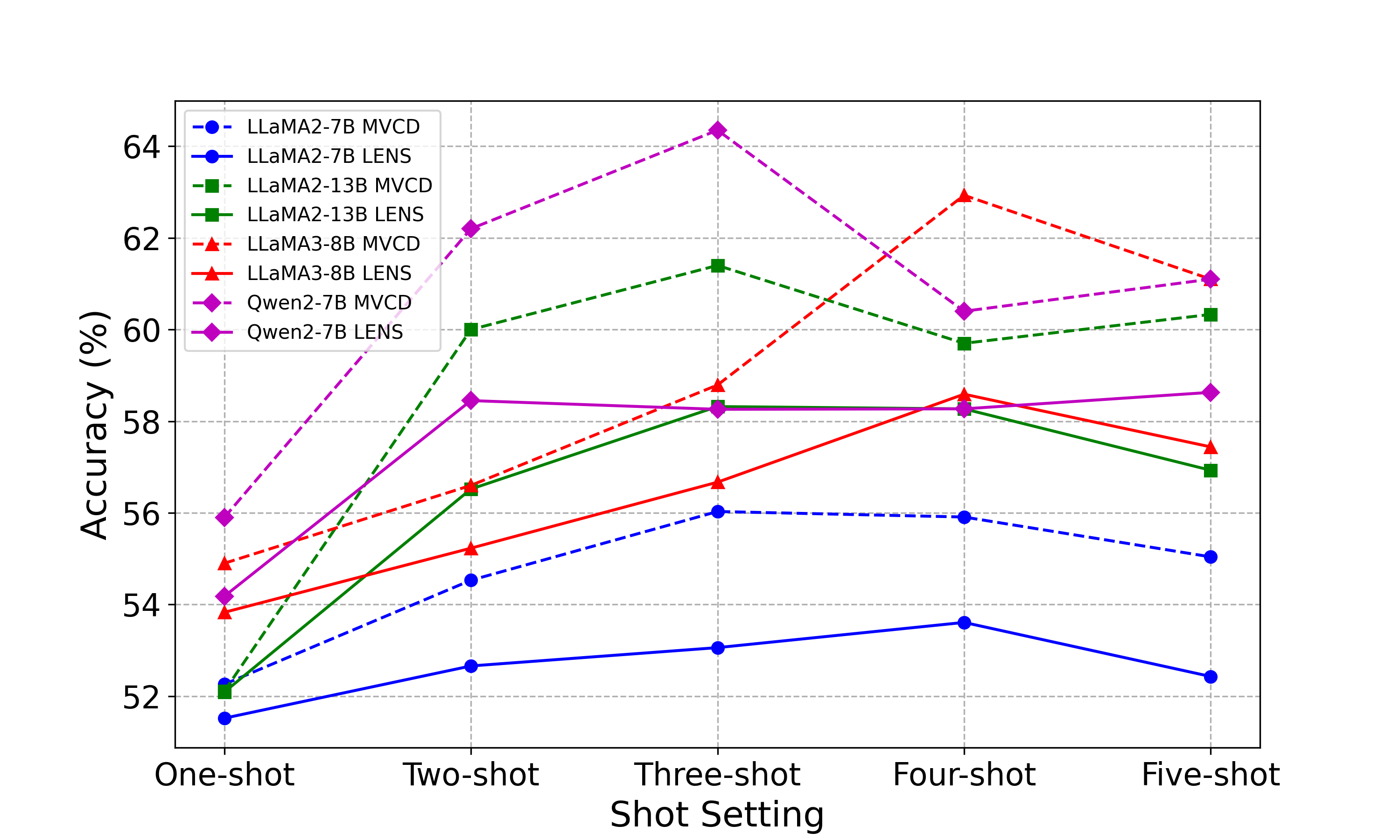}}
  \centerline{(a)}\medskip
\end{minipage}
\hfill
\begin{minipage}[b]{.46\linewidth}
  \centering
  \centerline{\includegraphics[width=1.1\linewidth]{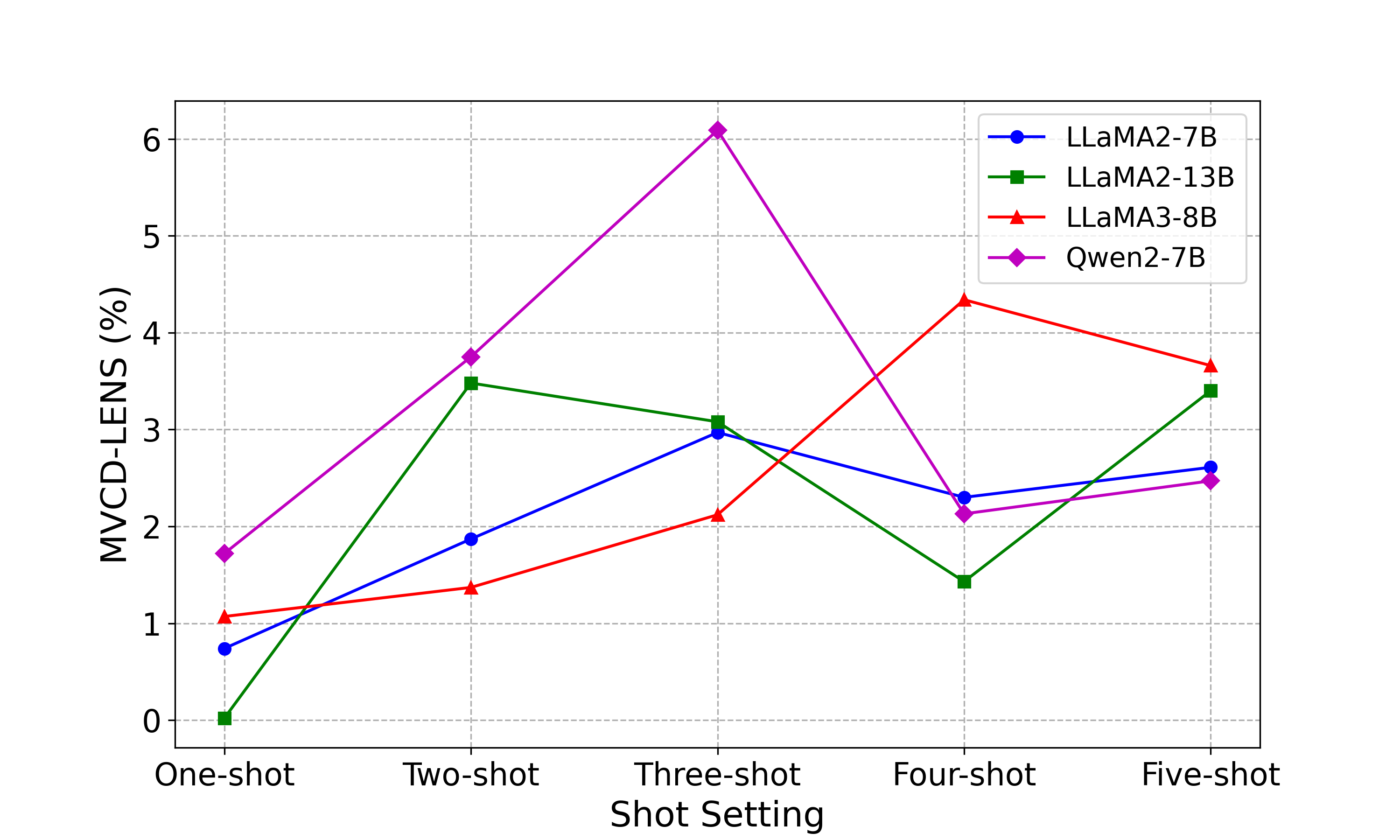}}
  \centerline{(b)}\medskip
\end{minipage}
\vspace{-1em}
\caption{(a) Evaluation results of MVCD and LENS on VQAv2, and (b) the performance gap between them on VQAv2.}
\label{fig:pe}
\end{figure}

However, the original output probability can still inherently capture much simple contextual information (e.g., grammar and common sense) and maintain strong relevance to the context. Thus, penalizing all behaviors from the original output distribution indiscriminately would penalize these simple aspects that are correct (i.e., False negative), and conversely reward implausible tokens (i.e., False positive). To address this, we introduce adaptive constraints that use the optimized output probability to limit the contrastive objective when the original output probability is high:

\begin{gather}
V_{head}(t_j|(E^1,E^2,\cdots,E^k),F,T_{<j}) \nonumber \\
=\{t_j \in V:\tilde{p}(t_j)\geq \alpha \mathop{\text{max}}\limits_{w}\tilde{p}(w) \},
\label{eq5}
\end{gather}

\begin{figure}[t]
\centerline{\includegraphics[width=\linewidth]{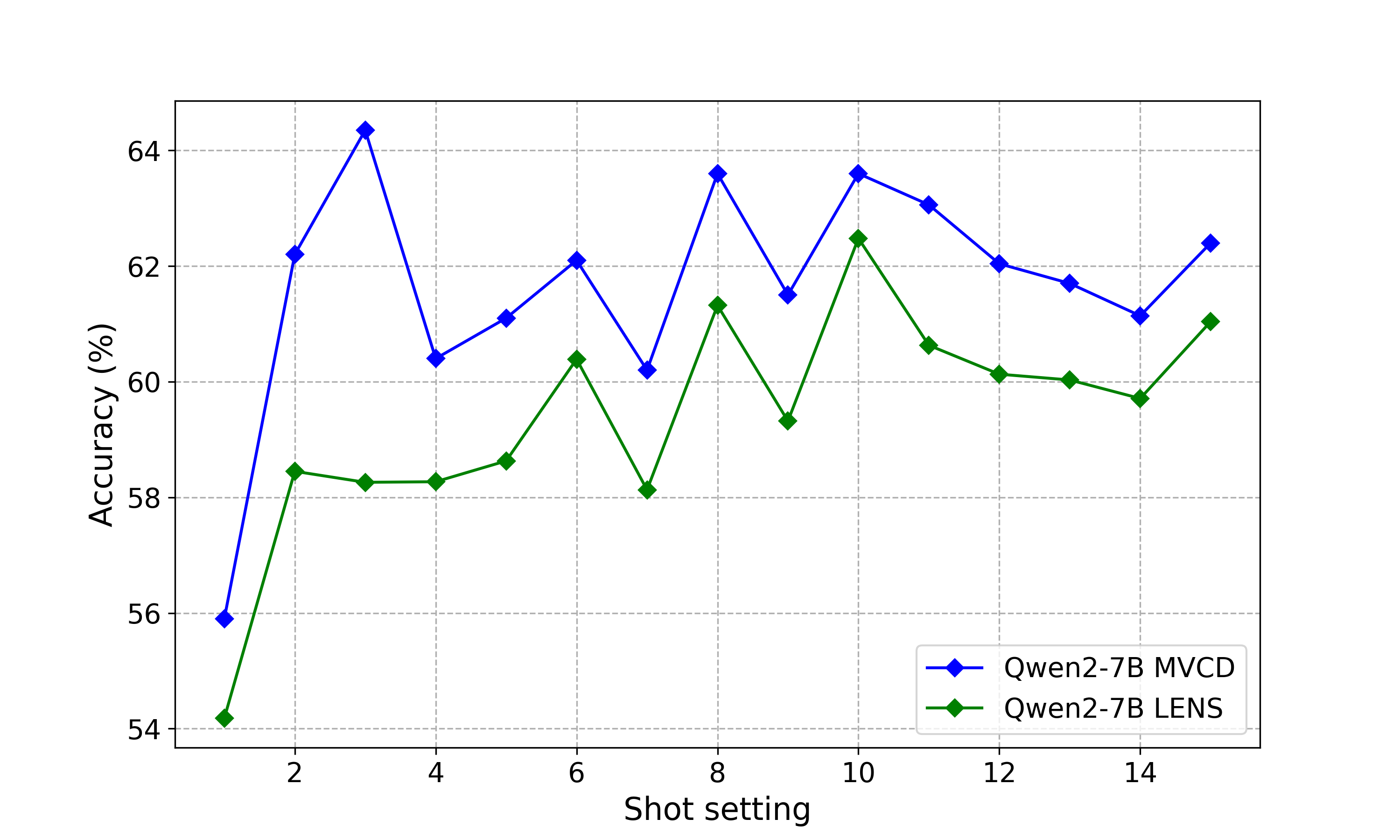}}
\caption{Ablation of input ICL shots: Results of MVCD and LENS on VQAv2 using Qwen2-7B with different ICL shot settings.}
\label{fig:shot}
\end{figure}
Here, $w$ is a token of the vocabulary $V$. To ensure coherence and consistency while avoiding repetition, we dynamically select an appropriate token based on \eqref{eq4} and \eqref{eq5}:
	\begin{equation}
	CED_{score}(t_j) = \begin{cases}
	\text{log} \frac{\tilde{p}(t_j)}{p(t_j)} &t_j\in V_{head},\\
	\text{-inf} & \text{otherwise}.	
    \label{eq7}
		   \end{cases}
\end{equation}
Therefore, we ultimately obtain the next token $t_j$ under the guidance of contextual examples information.
\section{Experiments}
\paragraph{Implementation Details} In \eqref{eq5}, $\alpha$ is a hyperparameter in $[0,1]$ that truncates the next token distribution of $\tilde{p}$.
Larger $\alpha$ entails more aggressive truncation, retaining only high-probability tokens, whereas smaller $\alpha$ allows the generation of lower-probability tokens. 
We set $\alpha=0.1$ throughout the paper. We set the default number of tags, attributes, and captions for each image or video to $N=5$. These experiments were conducted on 4 NVIDIA A40 (48GB) GPUs.
\paragraph{Models and Baselines} We examined our framework (with CED) using four LLMs:  LLaMA2-7B-Chat, LLaMA2-13B-Chat, LLaMA3-8B-Instruct, and Qwen2-7B, and compared them with two baselines: 1) original decoding (LENS), where different shot information can be added to the input based on the original LENS;
2) Dola \cite{r22} contrasts the probability
 differences between projections onto the vocabulary space at
 earlier and later layers, without using any visual input. 
\paragraph{Datasets}We evaluated our approach using five visual question answering datasets, including three image-question tasks (i.e., VQAv2 \cite{v2}, OKVQA \cite{okvqa}, MME \cite{r20}) and two video-question tasks (i.e., MSRVTT-QA \cite{vtt}, MSVD-QA). 
\paragraph{Comparison with MLLMs}We separately select MLLMs for image and video question answering tasks to comprehensively evaluate the differences
 and advantages of our approach. We conduct a quantitative
 assessment of video question-answering capabilities on two
 datasets and evaluate our approach for image understanding
 on three image question-answering benchmarks. Our results
 are based on the Qwen2-7B-Instruct model under three-shot
 and four-shot settings.

\section{Results and Analysis}


\subsection{Results Analyze}
Table \ref{tab:base} demonstrates that MVCD achieves competitive performance in both image and video question answering tasks, surpassing MLLMs in several key metrics. Notably, in video question-answering, this advantage can be attributed to the effective integration of temporal contextual information and the efficient application of a contrastive
example decoding strategy. 

In Table \ref{tab:22}, we observe a notable performance boost by adding contextual examples when
comparing LENS and Dola.
Specifically, LENS shows up to 9\% improvement when moving from zero to one example. 
CED, designed for our framework, outperforms Dola and LENS in the experiments, further validating its effectiveness within the MVCD framework. 
Thus, we focus on analyzing performance of MVCD (with CED) and draw the following conclusions:

\textbf{(i) MVCD enhances the visual perception capabilities of LLMs:} In Table \ref{tab:22}, compared with LENS and Dola, our MVCD shows overall performance improvements across the five datasets, as illustrated in Fig.~\ref{fig:pe}(a), with performance enhancements observed on VQAv2. As the number of examples increases, MVCD's performance boost is more significant than Dola's. Importantly, the original reasoning capabilities of LLMs are not weakened by the addition of the visual perception module; instead, they gain enhanced visual perception abilities. We speculate that the greater the performance gap between MVCD and LENS, the stronger its ability to learn associations between examples and contrastively enhance contextual knowledge. 

\textbf{(ii) More contextual examples do not necessarily lead to better performance: }In Table \ref{tab:22}, as the number of shots increases, the performance gap between MVCD and LENS reaches a saturation point and exhibits irregular fluctuations. For instance, Fig.~\ref{fig:pe}(b) shows that for LLaMA2-7B-Chat in the three-shot setting on VQAv2, the performance gap reaches saturation and then fluctuates. Too much contextual example information may negatively affect MVCD's performance.

\textbf{(iii) Stonger LLMs indicate better multimodal reasoning under the MVCD schema:} In Fig.~\ref{fig:pe}, LLaMA3-8B-Instruct and Qwen2-7B outperform LLaMA2-7B-Chat and LLaMA2-13B-Chat at the saturation point of the performance gap. This suggests that stronger LLMs perform better in multimodal reasoning tasks.

\begin{table}[t]
\centering
\renewcommand\arraystretch{0.8}
\caption{Ablation of examples selection strategy: results of MVCD using QWEN2-7B on VQAv2 with different input settings. }
\begin{tabular}{@{}l|ccccc@{}}
\toprule  
 Strategy & One-shot & Two-shot & Three-shot & Four-shot & Five-shot \\
\midrule
Random  & 53.71 & 58.56 &57.44 & 57.20& 55.74 \\
Question type& 55.90 & 62.20 & 64.35 & 60.40 & 61.10 \\
\bottomrule
\end{tabular}
\label{tab:stra}
\end{table}
\begin{table}[t]
\centering
\renewcommand\arraystretch{0.8}
\caption{Ablation of the visual module: Results of MVCD using Qwen2-7B on OKVQA with different input settings.}
\begin{tabular}{@{}l|cccccc@{}}
\toprule  
\multirow{2}{*}{Mehtod} & \multicolumn{5}{c}{Attributes + Captions} \\
\cmidrule(lr){2-7} & Zero & One & Two & Three & Four & Five \\
\midrule
LENS & 23.35 & 27.82 & 29.48 & 30.71 & 31.51 & 31.14\\
MVCD & - & +3.71 & +3.92 & +3.96 & +1.41 & +1.66 \\
\midrule
\multicolumn{7}{c}{Tags + Captions} \\  
\midrule
LENS & 23.20 & 28.66 & 29.88 & 30.77 & 31.98 & 32.80 \\
MVCD & - & +2.39 & +3.45 & +3.93 & +0.92 & +0.10 \\
\bottomrule
\end{tabular}
\label{tab:module}
\end{table}

\subsection{Ablation Studies}
In Fig.~\ref{fig:shot}, we present model results with varied number of ICL examples. Our method demonstrates better robustness than baselines since our performance is further boosted with increased number of examples. Table \ref{tab:stra} demonstrates that selecting examples based on question type is more effective than random selection, and example selection is crucial for achieving high performance in multimodal reasoning. In Table \ref{tab:module}, removing either the attribute or tag module results in a performance drop. In the baseline, the tag module is more valuable than the attribute module, as it provides useful visual information within a certain threshold. Conversely, the attribute module is more effective than the tag module for contrastively learning contextual examples knowledge. 
Therefore, the combination of tags, attributes, and captions is crucial for achieving high performance in this task. 

\section{Conclusion}
This paper introduces the MVCD framework with contrastive-example decoding (CED). Without task-specific training, it converts visual signals into text through a visual module. By using LLMs’ in-context learning (ICL) to compare output  probabilities with and without cross-modal examples, MVCD enhances multimodal reasoning and consistently performs well across five major datasets.

 The proposed framework exhibits distinct advantages. First,
 it fully leverages the strengths of multimodal large language
 models such as CLIP, LanguageBind, BLIP, and Video-LLaVA
 utilizing their pretrained outputs to achieve exceptional visual
 content understanding without requiring retraining. Second,
 these multimodal large language models can be seamlessly
 replaced with smaller models tailored to similar tasks, such
 as image recognition models or video captioning models. However, this paper does not include comparative
 experiments with different pretrained models, which will be
 explored in future research.

\bibliographystyle{IEEEtran}
\bibliography{main}

\begin{thebibliography}{10}
\providecommand{\url}[1]{#1}
\csname url@samestyle\endcsname
\providecommand{\newblock}{\relax}
\providecommand{\bibinfo}[2]{#2}
\providecommand{\BIBentrySTDinterwordspacing}{\spaceskip=0pt\relax}
\providecommand{\BIBentryALTinterwordstretchfactor}{4}
\providecommand{\BIBentryALTinterwordspacing}{\spaceskip=\fontdimen2\font plus
\BIBentryALTinterwordstretchfactor\fontdimen3\font minus \fontdimen4\font\relax}
\providecommand{\BIBforeignlanguage}[2]{{%
\expandafter\ifx\csname l@#1\endcsname\relax
\typeout{** WARNING: IEEEtran.bst: No hyphenation pattern has been}%
\typeout{** loaded for the language `#1'. Using the pattern for}%
\typeout{** the default language instead.}%
\else
\language=\csname l@#1\endcsname
\fi
#2}}
\providecommand{\BIBdecl}{\relax}
\BIBdecl

\bibitem{r4}
OpenAI, ``Chatgpt,'' \url{https://openai.com/blog/chatgpt/}, 2023.

\bibitem{r5}
S.~Zhang, S.~Roller, N.~Goyal, M.~Artetxe, M.~Chen, S.~Chen, C.~Dewan, M.~Diab, X.~Li, X.~V. Lin \emph{et~al.}, ``Opt: Open pre-trained transformer language models,'' \emph{arXiv preprint arXiv:2205.01068}, 2022.

\bibitem{r6}
H.~Touvron, T.~Lavril, G.~Izacard, X.~Martinet, M.-A. Lachaux, T.~Lacroix, B.~Rozi{\`e}re, N.~Goyal, E.~Hambro, F.~Azhar \emph{et~al.}, ``Llama: Open and efficient foundation language models,'' \emph{arXiv preprint arXiv:2302.13971}, 2023.

\bibitem{ic}
T.~B. Brown, ``Language models are few-shot learners,'' \emph{arXiv preprint arXiv:2005.14165}, 2020.

\bibitem{COT}
J.~Wei, X.~Wang, D.~Schuurmans, M.~Bosma, F.~Xia, E.~Chi, Q.~V. Le, D.~Zhou \emph{et~al.}, ``Chain-of-thought prompting elicits reasoning in large language models,'' \emph{Advances in neural information processing systems}, vol.~35, pp. 24\,824--24\,837, 2022.

\bibitem{liu2024visual}
H.~Liu, C.~Li, Q.~Wu, and Y.~J. Lee, ``Visual instruction tuning,'' \emph{Advances in neural information processing systems}, 2024.

\bibitem{tu2023sight}
H.~Tu, B.~Zhao, C.~Wei, and C.~Xie, ``Sight beyond text: Multi-modal training enhances llms in truthfulness and ethics,'' in \emph{NeurIPS 2023 Workshop on Instruction Tuning and Instruction Following}, 2023.

\bibitem{tu2023resee}
H.~Tu, Y.~Li, F.~Mi, and Z.~Yang, ``Resee: Responding through seeing fine-grained visual knowledge in open-domain dialogue,'' in \emph{EMNLP}, 2023.

\bibitem{r1}
J.~Bai, S.~Bai, S.~Yang, S.~Wang, S.~Tan, P.~Wang, J.~Lin, C.~Zhou, and J.~Zhou, ``Qwen-vl: A frontier large vision-language model with versatile abilities,'' \emph{arXiv preprint arXiv:2308.12966}, 2023.

\bibitem{r2}
W.~Dai, J.~Li, D.~Li, A.~M.~H. Tiong, J.~Zhao, W.~Wang, B.~Li, P.~Fung, and S.~Hoi, ``Instructblip: Towards general purpose vision-language models with instruction tuning,'' \emph{arXiv preprint arXiv:2305.0650}, 2023.

\bibitem{r3}
J.-B. Alayrac, J.~Donahue, P.~Luc, A.~Miech, I.~Barr, Y.~Hasson, K.~Lenc, A.~Mensch, K.~Millican, M.~Reynolds \emph{et~al.}, ``Flamingo: a visual language model for few-shot learning,'' \emph{Advances in neural information processing systems}, vol.~35, pp. 23\,716--23\,736, 2022.

\bibitem{zhang2023video}
H.~Zhang, X.~Li, and L.~Bing, ``Video-llama: An instruction-tuned audio-visual language model for video understanding,'' \emph{arXiv preprint arXiv:2306.02858}, 2023.

\bibitem{su2022language}
Y.~Su, T.~Lan, Y.~Liu, F.~Liu, D.~Yogatama, Y.~Wang, L.~Kong, and N.~Collier, ``Language models can see: Plugging visual controls in text generation,'' \emph{arXiv preprint arXiv:2205.02655}, 2022.

\bibitem{tu2023zerogen}
H.~Tu, B.~Yang, and X.~Zhao, ``Zerogen: Zero-shot multimodal controllable text generation with multiple oracles,'' in \emph{CCF International Conference on Natural Language Processing and Chinese Computing}, 2023.

\bibitem{r9}
W.~Berrios, G.~Mittal, T.~Thrush, D.~Kiela, and A.~Singh, ``Towards language models that can see: Computer vision through the lens of natural language,'' \emph{arXiv preprint arXiv:2306.16410}, 2023.

\bibitem{wei2024diffusion}
C.~Wei, C.~Liu, S.~Qiao, Z.~Zhang, A.~Yuille, and J.~Yu, ``De-diffusion makes text a strong cross-modal interface,'' in \emph{CVPR}, 2024.

\bibitem{tu2024many}
H.~Tu, C.~Cui, Z.~Wang, Y.~Zhou, B.~Zhao, J.~Han, W.~Zhou, H.~Yao, and C.~Xie, ``How many unicorns are in this image? a safety evaluation benchmark for vision llms,'' in \emph{ECCV}, 2024.

\bibitem{leevhelm}
T.~Lee, H.~Tu, C.~H. Wong, W.~Zheng, Y.~Zhou, Y.~Mai, J.~S. Roberts, M.~Yasunaga, H.~Yao, C.~Xie \emph{et~al.}, ``Vhelm: A holistic evaluation of vision language models,'' in \emph{The Thirty-eight Conference on Neural Information Processing Systems Datasets and Benchmarks Track}, 2024.

\bibitem{r7}
Y.~Su, T.~Lan, Y.~Wang, D.~Yogatama, L.~Kong, and N.~Collier, ``A contrastive framework for neural text generation,'' \emph{Advances in Neural Information Processing Systems}, vol.~35, pp. 21\,548--21\,561, 2022.

\bibitem{r22}
Y.-S. Chuang, Y.~Xie, H.~Luo, Y.~Kim, J.~Glass, and P.~He, ``Dola: Decoding by contrasting layers improves factuality in large language models,'' \emph{arXiv preprint arXiv:2309.03883}, 2023.

\bibitem{r10}
A.~Radford, J.~W. Kim, C.~Hallacy, A.~Ramesh, G.~Goh, S.~Agarwal, G.~Sastry, A.~Askell, P.~Mishkin, J.~Clark \emph{et~al.}, ``Learning transferable visual models from natural language supervision,'' in \emph{International conference on machine learning}.\hskip 1em plus 0.5em minus 0.4em\relax PMLR, 2021, pp. 8748--8763.

\bibitem{r11}
S.~Menon and C.~Vondrick, ``Visual classification via description from large language models,'' \emph{arXiv preprint arXiv:2210.07183}, 2022.

\bibitem{r12}
J.~Li, D.~Li, C.~Xiong, and S.~Hoi, ``Blip: Bootstrapping language-image pre-training for unified vision-language understanding and generation,'' in \emph{International conference on machine learning}.\hskip 1em plus 0.5em minus 0.4em\relax PMLR, 2022, pp. 12\,888--12\,900.

\bibitem{r13}
A.~Fan, M.~Lewis, and Y.~Dauphin, ``Hierarchical neural story generation,'' \emph{arXiv preprint arXiv:1805.04833}, 2018.

\bibitem{r14}
B.~Zhu, B.~Lin, M.~Ning, Y.~Yan, J.~Cui, H.~Wang, Y.~Pang, W.~Jiang, J.~Zhang, Z.~Li \emph{et~al.}, ``Languagebind: Extending video-language pretraining to n-modality by language-based semantic alignment,'' \emph{arXiv preprint arXiv:2310.01852}, 2023.

\bibitem{r15}
B.~Lin, B.~Zhu, Y.~Ye, M.~Ning, P.~Jin, and L.~Yuan, ``Video-llava: Learning united visual representation by alignment before projection,'' \emph{arXiv preprint arXiv:2311.10122}, 2023.

\bibitem{li2023blip}
J.~Li, D.~Li, S.~Savarese, and S.~Hoi, ``Blip-2: Bootstrapping language-image pre-training with frozen image encoders and large language models,'' in \emph{International conference on machine learning}.\hskip 1em plus 0.5em minus 0.4em\relax PMLR, 2023, pp. 19\,730--19\,742.

\bibitem{zhang2023llama}
R.~Zhang, J.~Han, C.~Liu, P.~Gao, A.~Zhou, X.~Hu, S.~Yan, P.~Lu, H.~Li, and Y.~Qiao, ``Llama-adapter: Efficient fine-tuning of language models with zero-init attention,'' \emph{arXiv preprint arXiv:2303.16199}, 2023.

\bibitem{maaz2023video}
M.~Maaz, H.~Rasheed, S.~Khan, and F.~S. Khan, ``Video-chatgpt: Towards detailed video understanding via large vision and language models,'' \emph{arXiv preprint arXiv:2306.05424}, 2023.

\bibitem{v2}
Y.~Goyal, T.~Khot, D.~Summers-Stay, D.~Batra, and D.~Parikh, ``Making the v in vqa matter: Elevating the role of image understanding in visual question answering,'' in \emph{Proceedings of the IEEE conference on computer vision and pattern recognition}, 2017, pp. 6904--6913.

\bibitem{okvqa}
K.~Marino, M.~Rastegari, A.~Farhadi, and R.~Mottaghi, ``Ok-vqa: A visual question answering benchmark requiring external knowledge,'' in \emph{Proceedings of the IEEE/cvf conference on computer vision and pattern recognition}, 2019, pp. 3195--3204.

\bibitem{r20}
C.~Fu, P.~Chen, Y.~Shen, Y.~Qin, M.~Zhang, X.~Lin, J.~Yang, X.~Zheng, K.~Li, X.~Sun, Y.~Wu, and R.~Ji, ``Mme: A comprehensive evaluation benchmark for multimodal large language models,'' \emph{arXiv preprint arXiv:2306.13394}, 2023.

\bibitem{vtt}
D.~Xu, Z.~Zhao, J.~Xiao, F.~Wu, H.~Zhang, X.~He, and Y.~Zhuang, ``Video question answering via gradually refined attention over appearance and motion,'' in \emph{Proceedings of the 25th ACM international conference on Multimedia}, 2017, pp. 1645--1653.

\end{thebibliography}

\end{document}